\documentclass[rmp,aps,onecolumn,nofootinbib]{revtex4}

\usepackage{amsmath}
\usepackage{tabularx}
\usepackage{palatino} 
\usepackage[]{graphicx}
\usepackage[]{setspace} 

\usepackage{epsfig}
\usepackage{epstopdf}

\begin{document}
\begin{spacing}{1.5}

\title{Notes on Generalized Linear Models of Neurons}
\date{\today, version 1.51}
\author{Jonathon Shlens} 
\email{jonathon.shlens@gmail.com}
\affiliation{
Google Research \\
Mountain View, CA 94043}
\begin{abstract}
Experimental neuroscience increasingly requires tractable models for analyzing and predicting the behavior of neurons and networks. The generalized linear model (GLM) is an increasingly popular statistical framework for analyzing neural data that is flexible, exhibits rich dynamic behavior and is computationally tractable~\cite{Truccolo2005, Paninski2004b, Pillow2008}. What follows is a brief summary of the primary equations governing the application of GLM's to spike trains with a few sentences linking this work to the larger statistical literature. Latter sections include extensions of a basic GLM to model spatio-temporal receptive fields as well as network activity in an arbitrary numbers of neurons.
\end{abstract}
\maketitle

The generalized linear model (GLM) is a powerful framework for modeling statistical relationships in complex data sets. One application of the GLM is to relate the activity of a neuron (or network) to a sensory stimulus and surrounding network dynamics. A GLM of a neuron requires a probabilistic model of spiking activity paired with a function relating extrinsic factors (e.g. preceding stimulus, refractory dynamics, network activity) to a neuron's internal drive. In what follows, the former is provided by a {\it Poisson likelihood} while the latter is provided by a (cleverly constructed) {\it conditional intensity}. Finally we discuss extensions of the GLM for modeling spatio-temporal receptive fields and correlated network activity.\footnote{This manuscript is based on notes by Jonathan Pillow and Liam Paninski and conversations with Eero Simoncelli and EJ Chichilnisky. The manuscript assumes that the reader is familiar with sensory physiology and has a working knowledge of probability, linear algebra and calculus. Bold letters $\mathbf{k}, \mathbf{h}, \mathbf{s}$ are column vectors while capitalized bold letters $\mathrm{X}$ are matrices.}

\section{Poisson likelihood}
\label{sec:poisson}
A common starting point for modeling a neuron is the idealization of a Poisson process~\cite{Dayan2001}. If $y$ is the observed number of spikes, the probability of such an event is given by a Poisson distribution
\begin{equation}P(y | \lambda) = \frac{(\lambda \Delta)^y}{y!} \exp(-\lambda \Delta)\end{equation}
where $\lambda \Delta$ is the expected number of spikes in a small unit of time $\Delta$ and $\lambda$ is the intensity of the Poisson process.
\addtolength{\parskip}{\baselineskip}  

A spike train $Y$ is defined as a vector of spike counts $\{y_t\}$ binned at a brief time resolution $\Delta$ and indexed by time $t$. The likelihood that the spike train arises from a time-varying (inhomogenous) Poisson process  with a conditional (time-varying) intensity $\lambda_t$ is the product of independent observations,
$$P(Y | \theta ) = \prod_t \frac{(\lambda_t \Delta)^{y_t}}{y_{t}!} \exp(-\lambda_t \Delta)$$
where $\theta = \{ \lambda_t\}$ refers to the collection of intensities over the entire spike train. The log-likelihood of observing the entire spike train is
$$ \log P( Y | \theta ) = \sum_t y_t \log \lambda_t + \sum_t y_t \log \Delta - \sum_t \log y_{t}! - \Delta \sum_t \lambda_t $$
In a statistical setting our goal is to infer the hidden variables $\theta$ from the observed spike train $Y$. Many criterion exist for judging the quality of values for $\theta$~\cite{Kay1993, Moon1999}. Selecting a particular set of values for $\theta$ that maximize the likelihood (or log-likelihood) is one such criterion that is intuitive and and relatively easy to calculate. The goal of this manuscript is to calculate the {\it maximum likelihood} estimate of hidden variables by optimizing (or maximizing) the log-likelihood. 

In our setting, the spike train has been observed and is therefore {\em fixed}. The log-likelihood is solely a function of the (unknown) conditional intensities $\theta$. To emphasize our application of likelihood, we relabel $\mathcal{L}(\theta) \equiv \log P( Y | \theta )$ and group the terms independent of $\theta$ as some arbitrary normalization constant $c$,
$$\mathcal{L}(\theta) = \sum_t y_t \log \lambda_t - \Delta \sum_t \lambda_t + c.$$
In practice, models of neurons are at fine temporal resolutions, thus $y_t$ is effectively binary, if not largely sparse. The binary representation of $y_{t}$ simplifies the log-likelihood further
\begin{equation}
\label{eqn:poisson}
\mathcal{L}(\theta) = \sum_{t = spike} \log \lambda_t - \Delta \sum_t \lambda_t.
\end{equation}
In the final form of the log-likelihood, we ignore the normalization constant $c$ because the goal of the optimization is to determine the hidden variables $\theta$ (not the absolute value of the likelihood).
\addtolength{\parskip}{-\baselineskip}  

\section{Generalized Linear Model}
\label{sec:glm}
\begin{figure*}[tbp] 
\centering
\includegraphics[width=0.85\textwidth]{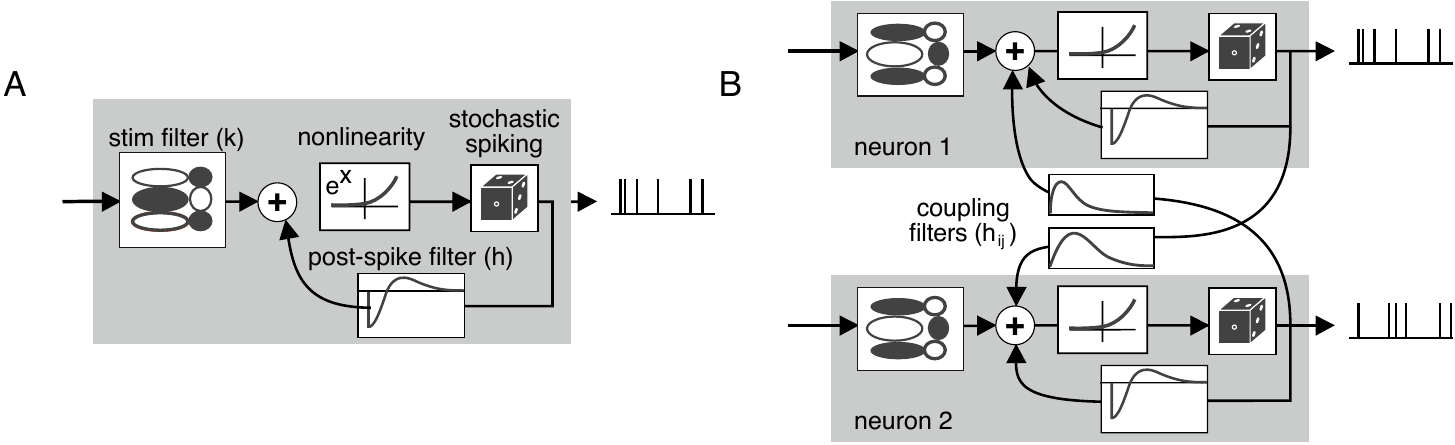}
\caption{Diagram of a generalized linear model of a single neuron (a) and a network of neurons (b).}
\label{fig:diagram}
\end{figure*}

A GLM can attribute the variability of a spike train to a rich number of factors including the stimulus, spike history and network dynamics. The GLM extends the maximum likelihood procedure to more interesting variables than $\lambda_t$ by positing that the conditional intensity is directly related to  biophysically interesting parameters. For example, pretend there is some parameter $\mathbf{k}$, a neuron's receptive field. If we posit that the receptive field is linearly related to some known (and invertible) function of $\lambda_t$, then in principle it is just as easy to infer $\mathbf{k}$ as it is to infer $\lambda_t$. By positing a linear relationship, the stimulus filter is quite simple to estimate, but surprisingly provides a rich structure for modeling the neural response.
\addtolength{\parskip}{\baselineskip}  

More precisely, the goal of a single neuron GLM is to predict the current number of spikes $y_t$ using the recent spiking history and the preceding stimulus. Let $\mathbf{x_t} = (x_{t-\tau}, \ldots, x_{t-1})$ represent the vector of preceding stimuli up to but not including time $t$. Let $\mathbf{y_t} = (y_{t-\tau}, \ldots, y_{t-1})$ be a vector of preceding spike counts up to but not including time $t$.

\begin{centering}
\includegraphics[width=0.3\textwidth]{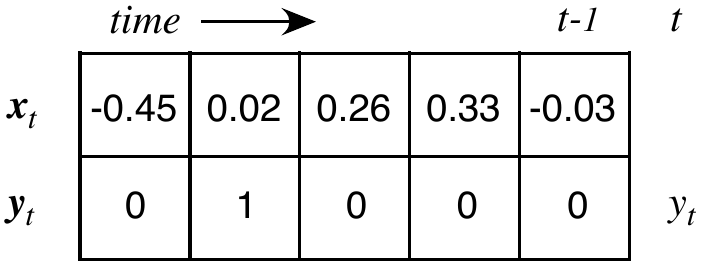}\\
\end{centering}
Importantly, note that $\mathbf{x_t}$ and $\mathbf{y_t}$ are fixed known variables but $y_t$ is an unknown random variable. We posit that $y_t$ is distributed according to a Poisson distribution whose conditional intensity $\lambda_t$ is related to the stimulus and previous spiking history by
\begin{equation}
\label{eqn:intensity}
\lambda_t = f(\mathbf{k} \cdot \mathbf{x}_{t} + \mathbf{h} \cdot \mathbf{y}_{t} + \mu)
\end{equation}
where $\mathbf{k}$ is a stimulus filter of the neuron (i.e. receptive field), $\mathbf{h}$ is a post-spike filter to account for spike history dynamics (e.g. refractoriness, bursting, etc.) and $\mu$ is a constant bias to match the neuron's firing rate. Figure~\ref{fig:diagram}a provides a diagram of Equation~\ref{eqn:intensity}. Each term of the conditional intensity increases or decreases the average firing rate depending on the preceding stimulus (term 1) and the spiking history of the neuron (term 2). $f$ is an arbitrary invertible function termed a {\em link function}. One possibilty is to select $f = \exp(\cdot)$, termed the {\em natural} link function, because it conveniently simplifies the likelihood calculation and the interpretation of individual terms~\cite{McCullagh1989}. In this case, a stimulus preceding time $t$ that closely matches the filter $\mathbf{k}$ increases the average spike rate by a multiplicative gain $\exp(\mathbf{k} \cdot \mathbf{x}_t)$. Likewise, if a spike occurs just prior to time $t$, the convolution of the spike occurrence (a delta function) with the post-spike filter diagrammed in Figure~\ref{fig:diagram}a decreases the probability of a spike by factor $\exp(\mathbf{h} \cdot\mathbf{y}_t)$ to mimic a refractory period.

It is worth noting at this point that the name {\em generalized linear model} refers to the requirement that $f^{-1}(\lambda)$ is linearly related to the parameters of interest, $\theta =\{\mathbf{k}, \mathbf{h}, \mu\}$.
Combining equations \ref{eqn:poisson} and \ref{eqn:intensity}, the probability of observing the complete spike train $Y$ is 
\begin{equation}\mathcal{L}(\theta)= \sum_{t=spike} (\mathbf{k} \cdot \mathbf{x}_{t} + \mathbf{h} \cdot \mathbf{y}_{t} + \mu) - \Delta \sum_t\exp(\mathbf{k} \cdot \mathbf{x}_{t} + \mathbf{h} \cdot \mathbf{y}_{t} + \mu)\end{equation}
Importantly, the likelihood is concave everywhere in the parameter space $\theta = \{\mathbf{k}, \mathbf{h}, \mu\}$, thus no local maxima exist and ascending the gradient leads to a single, unique global maximum. To calculate the maximum likelihood estimate of $\theta$ one must calculate the gradient and Hessian of the likelihood with respect to each variable,
\begin{center}
\begin{tabular}{|rcl|rcl|rcl|}
\multicolumn{3}{c}{Gradient} & \multicolumn{6}{c}{Hessian} \\
\hline
$\frac{\partial}{\partial \mathbf{k}} \mathcal{L}(\theta)$ & = & $\sum_{t = spike} \mathbf{x}_t - \Delta \sum_t \mathbf{x}_t \lambda_t$ & $\frac{\partial^2}{\partial \mathbf{k}^2} \mathcal{L}(\theta)$ & = & $- \Delta \sum_t \mathbf{x}_t \mathbf{x}_t^T \lambda_t$ & $\frac{\partial^2}{\partial \mathbf{k}\partial \mu} \mathcal{L}(\theta)$ & = & $- \Delta \sum_t \mathbf{x}_t\lambda_t$ \\
$\frac{\partial}{\partial \mathbf{h}} \mathcal{L}(\theta)$ & = & $\sum_{t = spike} \mathbf{y}_t - \Delta \sum_t \mathbf{y}_t \lambda_t$ & $\frac{\partial^2}{\partial \mathbf{h}^2} \mathcal{L}(\theta)$ & = & $- \Delta \sum_t \mathbf{y}_t \mathbf{y}_t^T \lambda_t$ & $\frac{\partial^2}{\partial \mathbf{h}\partial \mu} \mathcal{L}(\theta)$ & = & $- \Delta \sum_t \mathbf{y}_t\lambda_t$ \\
$\frac{\partial}{\partial \mu} \mathcal{L}(\theta)$ & = & $n_{sp}- \Delta \sum_t \lambda_t$ & $\frac{\partial^2}{\partial \mu^2} \mathcal{L}(\theta)$ & = & $- \Delta \sum_t \lambda_t$ & $\frac{\partial^2}{\partial \mathbf{k}\partial\mathbf{h}} \mathcal{L}(\theta)$ & = & $- \Delta \sum_t \mathbf{x}_t \mathbf{y}_t^T \lambda_t$ \\
\hline
\end{tabular}
\end{center}
where $n_{sp}$ denotes the total number of spikes in the spike train. These series of equations are solved by standard gradient-ascent algorithms (e.g. Newton-Raphson), permitting an estimate of the model parameters for a given spike train and stimulus movie. In practice, a model with 30 parameters can be fit with roughly 200 spikes.
\addtolength{\parskip}{-\baselineskip}  

\section{Receptive fields in space and time}
\label{sec:space-time}
Generalizing the GLM to multiple spatial dimensions, the conditional intensity retains the same form\footnote{For the remainder of this manuscript we drop the subscript $t$ on $\lambda_t$, $\mathbf{x}_t$ and $\mathbf{y}_t$ for notational simplicity.}  
\addtolength{\parskip}{\baselineskip}  
\begin{equation}\lambda = \exp( \sum_{i}^{s} \mathbf{k}_{i} \cdot \mathbf{x}_{i} + \mathbf{h} \cdot \mathbf{y} + \mu)\end{equation}
with the inner product of the stimulus filter and the stimulus summed over all $s$ spatial locations indexed by $i$. In practice though the number of parameters in the stimulus filter grows quadratically as the number of spatial locations and samples in the receptive field's integration period, making any estimate computationally slow and potentially intractable. To exploit matrix algebra in further equations, define $\mathrm{K} \equiv [\mathbf{k}_{1} \; \mathbf{k}_{2} \; \ldots \; \mathbf{k}_{s}]^T$ and $\mathrm{X} \equiv [\mathbf{x}_{1} \; \mathbf{x}_{2} \; \ldots \; \mathbf{x}_{s}]^T$. Rather than letting the number of parameters grow quadratically, one simplification is to posit that $\mathrm{K}$ is {\em space-time separable}, meaning that $\mathrm{K}$ can be written as an outer product,
$$\mathrm{K} = \mathbf{t}\mathbf{s}^T$$
where the vectors $\mathbf{s}$ and $\mathbf{t}$ independently capture the spatial and temporal components of the receptive field, respectively. Rewriting the conditional intensity with matrix algebra,
\begin{equation}
\lambda = \exp( \mathbf{s}^T \mathrm{X} \mathbf{t} + \mathbf{h} \cdot \mathbf{y} + \mu)
\end{equation}
the model is not linear, but quadratic, in parameter space $\theta = \{\mathbf{s}, \mathbf{t}, \mathbf{h}, \mu\}$ and the likelihood surface is not concave everywhere (i.e. local maxima exist). In particular, the lack of concavity arises from the new Hessian term measuring the dependency between the space and time filters.
\begin{equation}
\frac{\partial^2}{\partial \mathbf{s} \partial \mathbf{t}} \mathcal{L}(\theta) = \sum_{t=spike}{\mathrm{X}} - \Delta \sum_{t} \mathrm{X} \lambda_t - \Delta \sum_{t} \mathrm{X}\mathbf{t}\mathbf{s}^{T}\mathrm{X}\lambda_t
\end{equation}
Ascending the gradient, however, produces a good approximation of the standard model but with a notable reduction in the number of parameters. The gradient formulae are all the same as Section~\ref{sec:glm} if one replaces $\mathbf{x}_t$ with $\mathrm{X}\mathbf{t}$ for the derivatives associated with $\mathbf{s}$, and $\mathbf{s}\mathrm{X}$ for the derivatives associated with $\mathbf{t}$.

\addtolength{\parskip}{-\baselineskip}  

\section{Networks of neurons}
\label{sec:network}
The most significant feature of networks of neurons is that the activity of one neuron significantly influences the activity of surrounding neurons. One common example of correlated activity is the observation of synchronized firing, when two or more neurons fire nearly simultaneously more often then expected by chance. Importantly, correlated activity can be independent of shared stimulus drive and instead reflect an underlying biophysical mechanisms (e.g. common input, electrical or synaptic coupling, etc.).
\addtolength{\parskip}{\baselineskip}  

A naive implementation of the GLM (Equation~\ref{eqn:intensity}) for two neurons with conditional intensities $\lambda_1$ and $\lambda_2$ would fail to capture synchrony or any correlated activity because $\lambda_1$ and $\lambda_2$ are independent\footnote{More precisely, the neurons modeled by $\lambda_1$ and $\lambda_2$ would not be independent but {\em conditionally independent}~\cite{Schneidman2003a}. The latter distinction means that both neurons would be independent given that the stimulus is held fixed. This distinction is important because one could select a stimulus which drives both neurons and attribute the apparent correlated activity to an underlying mechanism as opposed to the choice of stimulus.}. A simple extension of the conditional intensity can mimic correlated activity such as synchrony~\cite{Pillow2008}. In particular, one can add post-spike filters $\mathbf{h}_{ij}$ to permit spikes from one neuron to influence the firing rate of another neuron (see Figure~\ref{fig:diagram}b). The conditional intensity of neuron $i$ is then
\begin{equation}
\lambda_i = \exp(\mathbf{k}_{i} \cdot \mathbf{x} + \sum_{j}\mathbf{h}_{ij} \cdot \mathbf{y}_j + \mu_i)
\end{equation}
where we have added the subscript $i$ throughout to label neuron $i$. The term $\sum_{j} \mathbf{h}_{ij} \mathbf{y}_j$ sums over post-spike activity received by neuron $i$, whether internal dynamics  ($j=i$) or activity from other neurons ($j\neq i$). 

The complete likelihood of a population of neurons is
\begin{equation}\mathcal{L}(\theta)= \sum_{i,t=spike}(\mathbf{k}_{i} \cdot \mathbf{x} + \sum_{j}\mathbf{h}_{ij} \cdot \mathbf{y}_j + \mu_i) - \Delta \sum_{i,t} \exp(\mathbf{k}_{i} \cdot \mathbf{x} + \sum_{j}\mathbf{h}_{ij} \cdot \mathbf{y}_j + \mu_i)\end{equation}
where $\theta = \{\mathbf{k}_i, \mathbf{h}_{ij}, \mu_i\}$ is the parameter set. Note again that the subscript for time $t$ is dropped from the stimulus and spike trains for clarity -- although the likelihood sums over this implicit subscript. The gradient of the likelihood of each post-spike coupling term $\mathbf{h}_{ij}$ must be generalized
$$\frac{\partial}{\partial \mathbf{h}_{ij}} \mathcal{L}(\theta) = \sum_{t = spike} \mathbf{y}_j - \Delta \sum_t \mathbf{y}_j \lambda_i$$
where care must be taken to pair the appropriate spike train $\mathbf{y}_j$ with conditional intensity $\lambda_i$. Most of the terms of the Hessian are preserved from the single neuron case (Section~\ref{sec:glm}) except for the terms associated with $\mathbf{h}_{ij}$:
$$\frac{\partial^2}{\partial \mathbf{h}_{ij}^2} \mathcal{L}(\theta) = - \Delta \sum_t  \mathbf{y}_j \mathbf{y}_j^T \lambda_i,\;\;\;\;  \frac{\partial^2}{\partial \mathbf{k}_i \partial \mathbf{h}_{ij}} \mathcal{L}(\theta) = - \Delta \sum_t  \mathbf{x}_i \mathbf{y}_j^T \lambda_i, \;\;\;\;\ \frac{\partial^2}{\partial \mu \partial \mathbf{h}_{ij}} \mathcal{L}(\theta) =- \Delta \sum_t  \mathbf{y}_j^T \lambda_i$$
Again, care must be taken to pair the appropriate spike train $\mathbf{y}_j $ with the appropriate conditional intensity $\lambda_i$. Note that Hessian terms between different neurons (e.g. $\mathbf{k}_i$, $\mathbf{k}_j$, for $j\neq i$) are zero due to the linearity of the conditional intensity. Because the cross terms of the Hessian between neurons are zero, one can fit a parameter set $\theta_i = \{\mathbf{k}_i, \mathbf{h}_{ij}, \mu_i\}$ for each neuron $i$ independently. In practice, this independence vastly simplifies the fitting procedure because each neuron can be fit in parallel.
\addtolength{\parskip}{-\baselineskip}  

\section{Conclusions}
This manuscript provided a very brief summary of analyzing neural data with generalized linear models (GLM's). The GLM is a flexible framework that allows one to attribute spiking activity to arbitrary phenomena -- whether network activity, sensory stimuli or other extrinsic factors -- and then estimate parameters efficiently using standard maximum likelihood techniques. Thus far, GLM's (or related ideas) have been used to model network activity in the retina~\cite{Pillow2008}, motor cortex~\cite{Truccolo2005}, visual cortex~\cite{Koepsell2008}, hippocampus~\cite{Brown2001} as well as devise new strategies for the design of experiments~\cite{Paninski2007}, quantify stimulus information in spike trains~\cite{Pillow2007} and help a lowly graduate student complete a Ph.D. thesis~\cite{Shlens2007}. 
\addtolength{\parskip}{\baselineskip}  

\bibliographystyle{apa}
\bibliography{glm}

\end{spacing}
\end{document}